\documentclass[11pt]{article}

\usepackage[preprint]{acl}
\usepackage{times}
\usepackage{latexsym}
\usepackage[T1]{fontenc}
\usepackage[utf8]{inputenc}
\usepackage{microtype}
\usepackage{inconsolata}
\usepackage{graphicx}
\usepackage{booktabs}
\usepackage{amsmath}
\usepackage{amssymb}
\usepackage{xcolor}
\usepackage{placeins}
\usepackage{threeparttable}

\title{Personalized Language Models for One-to-Many Private Communications}

\author{
  Zihan Hong \\
  The University of Queensland
  \And
  Xue Li \\
  The University of Queensland
}

\begin{document}
\maketitle

\begin{abstract}
We study \emph{relationship-specific continual reply prediction}
for one-to-many private communication: a system predicts an author's
reply to a recurring contact and updates its personalization state
when the actual reply becomes available. We propose \textbf{TRACE}
(Temporal Relationship-Aware Communication Evolution), which uses
author-level lexical habits, contact-specific wording, and temporal
expression estimates to guide a frozen language model, with a
feedback-driven policy controlling revision. On 216 test emails from
eight Enron users, all four main TRACE configurations achieve higher
mean BLEU, ROUGE-L, and BERTScore and lower mean Reply Style MSE than
Base LLM and Base RAG. Component ablations show mixed effects and do
not establish statistically reliable improvements in Reply Style MSE
after multiple-comparison correction. More accurate state forecasts
and larger revision budgets do not consistently improve reply-level
metrics. These exploratory findings expose the challenge of turning
evolving relationship-specific expression estimates into useful
generation guidance.

\noindent\textbf{Keywords:}
Personalized Text Generation; Large Language Models; Continual Personalization; Relationship-Aware Personalization; Email Reply Generation

\end{abstract}
\section{Introduction}
\label{sec:introduction}

A useful email assistant should help an author produce messages that reflect both their individual writing habits and the relationship with the recipient \citep{alhafni2024personalized,yao2026personamail}. Consider an author who initially answers a colleague's status requests with a greeting and a detailed confirmation. Over successive interactions, the author may shift toward brief acknowledgments as shared context accumulates, while continuing to provide fuller explanations to another contact. A reply can therefore be factually correct and fluent, yet still fail to reflect how the author now communicates with its recipient. An assistant that overlooks these changes may require the author to repeatedly correct wording and style that no longer fit the relationship.

Maintaining personalization over time requires responding to evolving expression and preferences \citep{jin2026twice,kim2026spring,liang2026pahf}. In private correspondence, this challenge arises across multiple relationships simultaneously. Some word-use patterns remain common across an author's writing, whereas others emerge or change within exchanges with a particular contact. These changes need not occur at the same time or follow the same direction across relationships. A dyadic expressive state captures the author's current wording and stylistic tendencies toward a specific contact, and its trajectory describes how these tendencies evolve over time. Updating a single author-level profile can incorrectly extend a change from one relationship to others, while retaining a fixed profile for each contact can overlook subsequent changes. The challenge is to continually incorporate new evidence while distinguishing shared habits from relationship-specific evolution.

These challenges motivate studying one-to-many private communication as a continual reply prediction problem. For each incoming message, the system must predict how the author would reply to that contact using the current message and the interactions observed so far. Once the author's actual reply becomes available, the system can use it to revise its estimate of the relevant dyadic state and guide future predictions. This process repeats as correspondence unfolds: each completed interaction provides evidence for subsequent adaptation, while changes within one relationship need not extend to others. We therefore ask: \emph{How can a system continually learn from newly observed interactions, update its estimates of an author's expressive states across recurring relationships, and use these evolving states to predict subsequent replies?}

To investigate this problem, we introduce \textbf{TRACE}
(\textbf{T}emporal \textbf{R}elationship-\textbf{A}ware
\textbf{C}ommunication \textbf{E}volution), which combines
lexical modeling, temporal estimation, and reinforcement learning.
TRACE models the author's general word-use patterns and
contact-specific preferences, and estimates their evolution
within each relationship. Following the use of historical
examples in personalized generation \citep{salemi2024lamp},
TRACE conditions a frozen language model on retrieved
interactions and these behavioral estimates to generate
personalized replies. A reinforcement learning policy then
decides whether and how to revise each reply. Both the behavioral
estimates and the revision policy are updated as observed replies
become available, informing subsequent interactions.

Our contributions are threefold.
First, we formulate relationship-specific continual reply
prediction for one-to-many private communication, where
observed human replies inform subsequent predictions across
an author's recurring contacts.
Second, we introduce TRACE, a framework that combines
author-level lexical habits, contact-specific preferences,
and temporal state estimates to guide a frozen language
model, together with a feedback-driven revision policy.
Third, we evaluate TRACE on an Enron email subset through
chronological comparisons with unpersonalized and retrieval-only
baselines, component ablations, and sensitivity analyses.
TRACE achieves better mean reply-level metrics than the two
baselines in this setting. However, the ablations do not establish
statistically reliable component-level improvements in Reply
Style MSE after multiple-comparison correction.
\section{Related Work}
\label{sec:related_work}

\paragraph{Personalization in Interpersonal Communication.}
Personalized generation examines how language reflects
an individual's identity and writing habits
\citep{li2016persona,zhang2018personachat,madotto2019personalizing}.
History-based benchmarks evaluate the role of prior writing
\citep{salemi2024lamp,kumar2024longlamp},
while personalized generation methods seek closer agreement
with individual expression
\citep{mysore2024pearl,tan2024oppu,
alhafni2024personalized,zhang2025stylevector}.
Communication introduces a further requirement:
expression also depends on the recipient.
Research on addressees, interlocutor-aware dialogue,
and email generation addresses this relational context
\citep{ouchi2016addressee,liu2019interlocutor,
chen2014email,gupta2018impersonation}.
Panza focuses on reproducing an author's writing
\citep{nicolicioiu2026panza}, while PersonaMail supports
relationship-sensitive email composition by reusing
communication strategies and learning style preferences
from user edits \citep{yao2026personamail}.
These studies motivate distinguishing shared author habits
from expression specific to a particular relationship.

\paragraph{Personalization Across Interactions.}
Recurring interactions require personalization to incorporate
new evidence over time.
Long-term dialogue research examines persistent memory,
temporal context, and recurring or changing partners
\citep{xu2022goldfish,zhong2024memorybank,
jang2023chronicles,jang2024mixed}.
Dynamic personalization benchmarks evaluate updated knowledge,
evolving profiles, and implicit preferences
\citep{wu2025longmemeval,jiang2025personamem,
jiang2025personamemv2}.
PersonaConvBench and REALTALK examine users' own messages
and longer-term interactions
\citep{li2025personaconvbench,lee2025realtalk}.
TWICE models evolving user expression in event-grounded
social-media post generation \citep{jin2026twice}.
SPRInG selectively updates user-specific adapters from
new interactions and combines them with retrieved history
\citep{kim2026spring}.
Other approaches learn from user-specific evidence and
feedback \citep{gao2024cipher,liang2026pahf,
salemi2024optimization,zhao2025rlpa}.

We study \emph{relationship-specific continual reply prediction}:
given an incoming message from a recurring contact, predict
the author's reply using only previously completed exchanges.
PersonaMail centers on interactive email composition and
feedback from user edits; TWICE generates posts conditioned
on events; and SPRInG adapts user-specific model parameters.
In TRACE, observed replies update author-wide lexical habits,
contact-specific lexical and temporal estimates, and an
external revision policy after they arrive, while the
language model remains frozen. We evaluate this setting
through chronological replay across an author's contacts.
\section{Problem Formulation}
\label{sec:problem_formulation}

We study continual personalized reply prediction in longitudinal, one-to-many private communication. An author $u$ maintains separate correspondence with a set of recurring contacts $\mathcal{C}_u$, where each author--contact pair $(u,c)$ defines a communication relationship. Each interaction $i$ consists of an incoming message $x_i$
from contact $c_i \in C_u$, with recorded timestamp $t_i$,
and the author's actual reply $y_i$, with recorded timestamp
$\tau_i \ge t_i$. The system predicts the reply at $t_i$
before observing $y_i$. When timestamps coincide, prediction
precedes reply feedback under the replay protocol.

The correspondence history available at prediction time is
\begin{equation}
\mathcal{H}_u^{<t_i}
=
\left\{
(c_j, x_j, y_j, t_j, \tau_j)
\mid \tau_j < t_i
\right\},
\end{equation}
where $j$ indexes interactions involving author $u$. For each interaction, $c_j$ denotes the contact, $x_j$ the
incoming message, and $y_j$ the author's actual reply;
$t_j$ and $\tau_j$ denote the recorded incoming-message
and reply timestamps, respectively. Thus, $\mathcal{H}_u^{<t_i}$ contains completed interactions across contacts whose replies are observable strictly before $t_i$. An earlier incoming message whose reply is not yet observable does not enter this completed-exchange history. The amount and timing of available evidence may vary across relationships.

Given $x_i$, $c_i$, $t_i$, and $\mathcal{H}_u^{<t_i}$, the task is to generate a reply $\hat{y}_i$ that approximates the author's actual reply $y_i$ in both content and expression. We use \emph{dyadic expressive state} to refer to the author's tendencies in wording and style when addressing a particular contact at a given time. The evolution of these tendencies over time constitutes that relationship's \emph{expressive trajectory}. These trajectories may differ across contacts: a change in how the author communicates with one contact need not occur in another relationship. The prediction target therefore depends on both the target contact and the author's expression within that relationship at the current time.

Predictions proceed chronologically. Once $y_i$ becomes observable at $\tau_i$, it provides a reference for evaluating $\hat{y}_i$, and the completed exchange becomes available as feedback for predictions made after $\tau_i$. As this process repeats, newly observed interactions provide evidence for revising the understanding of the author's expressive states across relationships. Only actual author-written replies enter the observed history; generated replies do not substitute for real interactions. The objective is to maintain accurate reply prediction over time by adapting to changes in the author's expression within each recurring relationship.
\section{TRACE Framework}
\label{sec:trace_framework}

\begin{figure*}[!t]
    \centering
    \includegraphics[width=\textwidth]{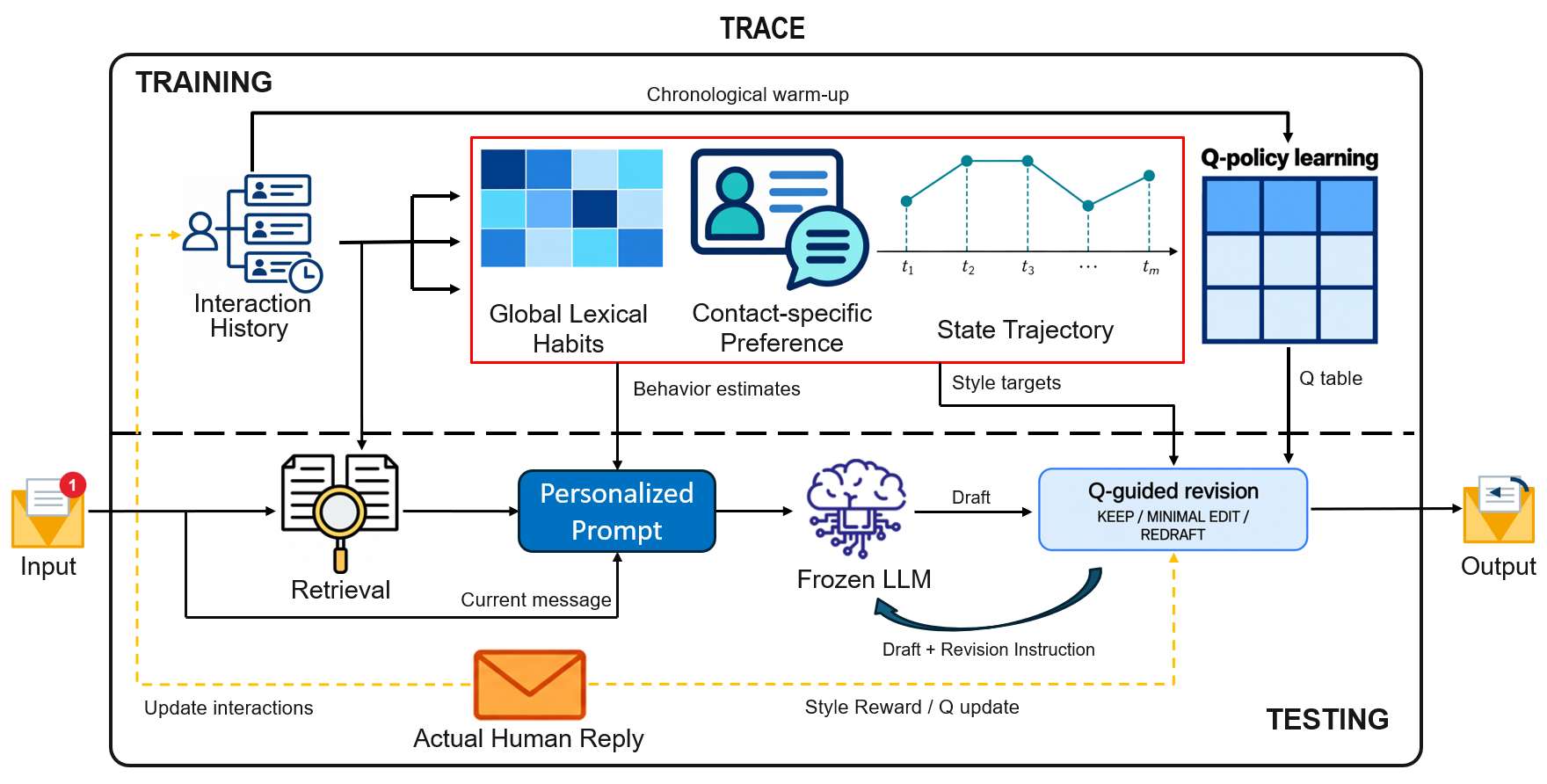}
    \caption{Overview of TRACE. Solid arrows indicate data and
    control flow; dashed arrows indicate updates triggered by the
    arrival of human replies.}
    \label{fig:trace_architecture}
\end{figure*}

Figure~\ref{fig:trace_architecture} summarizes TRACE. At query time $t$, all estimates use only
completed exchanges in $\mathcal{H}^{<t}_u$. The framework combines
author-level lexical habits, contact-specific wording, and temporal
estimates to construct a prompt for a frozen language model. A separate
Q-policy controls revision; observed human replies update the history
and policy only after their arrival.

\paragraph{Global and relationship-specific wording.}
The global profile contains the 40 most frequent words in the author's
observed replies, using pooled token counts. For the current contact,
we compare pooled relative word frequencies with the equally weighted
mean across the author's other observed contacts. We weight positive
differences by inverse contact frequency and retain the top 20 words.
This contrast identifies relationship-specific wording relative to the
same author's other correspondence. Here $m$ is the number of contacts observed before the query, and
$\mathrm{df}_u(w)$ counts those contacts whose replies contain $w$.
The inverse-contact-frequency weight is
$1+\log[(1+m)/(1+\mathrm{df}_u(w))]$.

\paragraph{Temporal expression estimates.}
Each historical reply is represented by its 20 normalized style
features, word-frequency distribution, supported ending-form
indicators, and signature/footer presence.
For a contact with sufficient history, we fit a slope-regularized linear
trend to the latest $W$ replies at their observed reply times. Let
$\bar{\mathbf v}$ denote the mean over all available replies to this
contact and $\tilde{\mathbf v}(t)$ the bounded trend extrapolation.
The forecast is
\begin{equation}
\widehat{\mathbf v}(t)=\mathcal{P}\!\left[
(1-\omega)\bar{\mathbf v}+\omega\tilde{\mathbf v}(t)
\right],
\label{eq:trace-forecast}
\end{equation}
where $\omega=0.5\exp(-g/3)$ and $g$ is the elapsed time since the latest
reply, measured in units of the contact's historical reply spacing.
The projection $\mathcal{P}$ clips bounded features and normalizes the
word distribution. The time unit is the larger of one day and the median positive gap
between consecutive replies in the contact's visible history;
extrapolation is capped at one such unit. Temporal fitting requires
at least $W$ replies and two distinct timestamps among the latest
$W$; otherwise, the forecast uses historical means. Without contact
history, it uses the author's available replies; with no history,
no forecast is supplied.

\paragraph{Generation and delayed revision feedback.}
The prompt combines these estimates, the current message, retrieved
exchanges, and summaries of the latest observed relationship states.
These are soft expression cues, not factual evidence for the current
reply. After the initial draft, the controller chooses
\textsc{Keep}, \textsc{Minimal Edit}, or \textsc{Redraft} using the
contact, decision index, and a discretized draft-to-forecast mismatch.
It cannot access the current human reply. Once that reply arrives,
the reduction in 20-feature Reply Style MSE supplies the reward for
updating the recorded decisions. The final draft is returned without
reference-based selection. When a forecast is available, revision prompts also supply
draft-to-forecast differences for the 11 controllable style features.

\section{Experiments}
\label{sec:experiments}

\subsection{Dataset}
\label{sec:dataset}
We construct our experimental subset from the Enron Email-Reply Dataset
\citep{ilie2025enronreply}, derived from the Enron corpus
\citep{klimt-yang-2004-enron}. We downloaded its pre-paired CSV from Kaggle
and applied study-specific data cleaning and selection. We normalized
participant addresses, parsed timestamps, and selected users with multiple
contacts and frequent correspondence to study repeated interactions.
We retained the source dataset's message--reply associations.
Data provenance and preprocessing are described in
Appendix~\ref{app:source-provenance}.

The resulting subset contains 1,020
incoming-message--reply pairs from eight target
users and 35 retained user--contact relationships,
covering 43 distinct email accounts.
Each target user has three to five retained
contacts, and the number of distinct reply dates
per relationship ranges from six to thirty.
Appendix~\ref{app:communication_network} visualizes
the retained relationships within the broader
Enron communication network, with additional
neighboring accounts shown only for context.

The retained message--reply pairs satisfy $\tau_i \ge t_i$.
Among the 1,020 pairs, 78 have identical recorded
incoming-message and reply timestamps. We retain these pairs
and apply the prediction-before-feedback ordering described
in Section~\ref{sec:experimental_setup}.

\subsection{Experimental Setup}
\label{sec:experimental_setup}
Within each user--contact relationship containing $n$ exchanges,
pairs are sorted by incoming-message timestamp, then reply timestamp,
and then case identifier, all in ascending order.
The first $\lfloor 0.8n \rfloor$ pairs form the training partition,
and the remaining pairs form the test partition.
This yields 804 training and 216 test examples.
Training and test interactions follow a single chronological
event stream per user; partition labels do not change feedback visibility.
Timestamps without timezone information are interpreted as UTC
for ordering; the original source timezone has not been
independently recovered.
The main experiment and ablation study use these same
216 test examples.
The sensitivity study uses a development subset of the
training partition, as described in
Section~\ref{sec:kw-sensitivity}.

All experiments follow a chronological online replay protocol.
A historical interaction is available for a prediction at
time $t$ only if its recorded reply timestamp is strictly
earlier than $t$. At each timestamp, all predictions are
completed before any reply feedback at that timestamp is
revealed. This ordering also applies when an incoming message
and its paired reply share the same timestamp. The current
reference reply is therefore excluded from its own prediction
and can affect only predictions at later timestamps.
Consequently, an earlier test reply may inform later
predictions after its feedback has been revealed. Observed human replies update the available histories and,
where enabled, the personalization profiles and revision policies.
Histories and learning states are isolated across users.

All methods use \texttt{gpt-5.6-terra} through the OpenAI
Responses API with low reasoning effort and
\texttt{store=false}.
No language-model fine-tuning is performed.
The API client does not explicitly set temperature,
top-$p$, an output-token limit, or a sampling seed.
Matching user-specific seeds control local policy sampling
across configurations; they do not seed API generation.

Cached responses are reused only when the case identity
and complete generation request match, including the model,
instructions, prompt, and generation settings.
Previously completed configurations retain their original
response bindings.
Reuse is independent of evaluation scores.
Each user--configuration pair contributes one online
trajectory; shared cached responses do not constitute
independent repeated runs.
We use $K$ to denote the maximum generation budget,
including the initial draft and up to $K-1$ subsequent
controller decisions.
For TRACE, $W$ denotes the number of recent completed
interactions with the current contact used for temporal fitting.

\subsection{Baselines}
\label{sec:baselines}

\textbf{Base LLM.}
The model generates a single reply from the current incoming
message and participant identities, without historical
interactions or explicit personalization summaries.
Its generation budget is $K=1$.

\textbf{Base RAG.}
Following the principle of generation conditioned on retrieved
evidence~\citep{lewis2020retrieval}, this baseline additionally
receives up to three completed historical exchanges.
It uses the same TF--IDF retriever as TRACE, with lowercase
unigram and bigram features and sublinear term-frequency scaling.
Retrieval uses the current contact's available history
when present and otherwise falls back to the same user's
available history.
The current query is excluded from fitting the retrieval index.
The index is fitted to historical incoming messages in the selected
pool, and exchanges are ranked by TF--IDF cosine similarity to
the current incoming message.
Similarity ties are broken by more recent reply timestamps
and then by descending case identifier.
Base RAG generates one reply with $K=1$ and does not construct
explicit lexical profiles or temporal forecasts,
learn a revision policy, or revise its output.

\subsection{TRACE Configurations}
\label{sec:trace}

TRACE conditions generation on retrieved exchanges,
the user's global lexical habits, contact-specific lexical
patterns, and temporal forecasts.
We vary the maximum generation budget $K$ and trajectory
window $W$.
The budget includes the initial draft and permits
$K-1$ subsequent controller decisions.
A \textsc{Keep} decision consumes a slot without making
another generation request, so the actual number of
generated replies can be smaller than $K$.
The window $W$ specifies the number of recent completed
interactions with the current contact used for temporal fitting.
It is distinct from the number of retrieved examples and
does not restrict the history used to estimate long-term
lexical habits.
Temporal fitting requires at least $W$ completed replies to the
current contact and at least two distinct reply timestamps among
the latest $W$ replies.
Otherwise, the forecast uses equally weighted historical means.
If no contact history is available, it uses the same user's
available replies; if no history is available, no forecast is supplied.

After the initial draft, the controller selects among
\textsc{Keep}, \textsc{MinimalEdit}, and \textsc{Redraft}.
Its state comprises the contact, decision index, and an 11-feature
mismatch bin: \texttt{low} for mean squared draft-to-forecast discrepancy
below $0.02$, \texttt{high} otherwise, and \texttt{no\_history} when no
forecast is available.
Each learning configuration starts with an empty
user-specific Q-table and uses $\epsilon$-greedy action
selection with $\epsilon=0.2$, learning rate $\alpha=0.2$,
discount factor $\gamma=1$, and uniform tie handling.
When the reference reply arrives, each transition receives
a reward equal to 100 times the reduction in Reply Style MSE
between its preceding and resulting draft states.
Multi-step updates proceed in reverse order, with zero
bootstrap at the terminal decision.
The final state is always evaluated; reference scores
are never used to select an intermediate draft.

The main experiment evaluates four configurations:
TRACE (reference), $(K,W)=(2,3)$;
TRACE (more revisions), $(3,3)$;
TRACE (short window), $(2,2)$;
and TRACE (long window), $(2,5)$.
All four retain global and contact-specific lexical modeling,
temporal forecasting, retrieval, and Q-learning.
Each configuration learns its own policy under its
corresponding decision horizon.
For a fixed query and $W$, changing $K$ does not alter
the pre-generation forecast because profile histories
contain observed replies rather than generated drafts.
Retrieval settings and generator options remain fixed,
while prompt lengths and generation budgets can differ.

\subsection{Ablation Setup}
\label{sec:ablation-setup}
We compare five targeted interventions against Full TRACE
($K=2$, $W=3$) on the same 216 test emails from eight users.
\emph{w/o Global Lexical Summary} removes the explicit global
word-frequency summary from both initial and revision prompts.
\emph{Shuffled Contact TF-IDF} replaces only the relationship
lexical contrast with that of another contact of the same user,
using a fixed permutation with no self-matches.
The true addressee, contact-history count, and contact-specific
forecasts remain unchanged.
\emph{w/o Temporal Modeling} replaces temporal projections of
style, lexical preferences, and ending forms with equally weighted
historical estimates, removing recency weighting, trend
extrapolation, and explicit trajectory cues from the profile.
\emph{w/o Q-learning} disables Q-value updates and samples
uniformly from \textsc{Keep}, \textsc{Minimal\_Edit}, and
\textsc{Redraft}, while retaining $K=2$ and Full TRACE's exact
initial drafts.
\emph{w/o Revision} returns those initial drafts directly ($K=1$).

The global-summary intervention retains the cross-contact
statistics used to compute the relationship lexical contrast.
The shuffled-contact intervention preserves retrieval and the
contact-indexed Q-state representation; the donor profile uses
only replies already available at the query time.
For the temporal intervention, static estimates are recomputed
from all causally available replies to the current contact,
retaining Full TRACE's fallback to the user's history when
contact history is unavailable.
The profile therefore continues to update as replies arrive.
Full TRACE's retrieved examples and their order are fixed for
every training and test query.
Thus, this intervention concerns temporal profile modeling;
causal history filtering, recency-based retrieval tie-breaking,
and chronological Q-learning feedback remain in place.
All variants share the generator settings and evaluation protocol.
Different \textsc{Keep} rates can change actual generation counts
even when $K$ is fixed.

The reference is the Full TRACE run used for this ablation study,
whose generated replies differ from those in the main experiment.
The temporal variant is replayed from an empty Q table for each
user, while the other five rows retain their completed-run outputs.
Saved responses are reused only for identical complete requests
for the same case, without selection based on evaluation scores.
Each row represents one online trajectory per user;
shared responses do not constitute independent repeated runs.

\subsection{Evaluation Metrics and Statistical Analysis}
\label{sec:evaluation}
We evaluate State Prediction MSE, BLEU, ROUGE-L,
BERTScore, and Reply Style MSE.
Both MSE metrics use the same 20 fixed normalized
style features (Appendix~\ref{app:trace-style-features}).
State Prediction MSE compares the state forecast made
before generation with the reference reply's feature vector.
Reply Style MSE compares the final generated reply
with the reference in the same feature space.
The historical-mean ablation supplies a valid state
forecast, whereas State Prediction MSE is not applicable
to Base LLM or RAG.

Before reply-level scoring, generated text is stripped of
leading or trailing whitespace, an optional leading
\texttt{Reply:} or \texttt{Response:} label, and matching
outer quotation marks.
Reference replies are scored as stored.
We compute corpus BLEU separately for each user using
SacreBLEU 2.6.0 with its default corpus scorer.
ROUGE-L is computed with \texttt{rouge-score} 0.1.2 and
stemming enabled.
ROUGE-L and BERTScore are example-level F1 scores,
and BERTScore uses \texttt{roberta-large}, layer 17,
without IDF weighting or baseline rescaling.
BLEU, ROUGE-L, and BERTScore are reported on a 0--100 scale.
For ROUGE-L, BERTScore, and both MSE metrics, we first
average scores within each user.
We then report the equally weighted mean of the eight
user-level scores and, where tabulated, their sample
standard deviation.
These standard deviations describe between-user variability,
not uncertainty across repeated generation runs.

Reply Style MSE is the primary outcome for the ablation study.
Each of the five ablations is compared with Full TRACE
using a two-sided paired sign-flip test over the eight
user-level score differences.
The tests enumerate all $2^8=256$ sign assignments and
assume symmetric paired differences under the null.
We apply Holm correction across the five
comparisons~\citep{holm1979simple}.
The remaining metric and configuration comparisons
are reported descriptively.
Given the small user sample, one online trajectory per
configuration, and prior inspection of the test set,
the findings are treated as exploratory.

\section{Results and Analysis}
\label{sec:results}

\subsection{Main Results}
Table~\ref{tab:trace-six-methods} compares Base LLM, RAG,
and four TRACE configurations on the same 216 test emails
from eight users.
All methods use GPT-5.6-terra and follow the same chronological
online evaluation protocol.
The reference TRACE configuration uses $K=2$ and $W=3$.

\begin{table*}[t]
    \centering
    \begin{threeparttable}

    \caption{
        Main results on 216 test emails from eight users using
        GPT-5.6-terra. Values are equally weighted user means
        $\pm$ between-user sample standard deviations,
        with one online run per user and method.
    }
    \label{tab:trace-six-methods}

    \small
    \setlength{\tabcolsep}{2pt}

    \begin{tabular}{lccrrrrr}
        \toprule
        Method
        & $K$
        & $W$
        & \shortstack{State Prediction\\MSE $\downarrow$}
        & BLEU $\uparrow$
        & ROUGE-L $\uparrow$
        & BERTScore $\uparrow$
        & \shortstack{Reply Style\\MSE $\downarrow$} \\
        \midrule

        Base LLM
        & 1
        & --
        & --
        & 0.785 $\pm$ 0.599
        & 9.282 $\pm$ 2.446
        & 84.188 $\pm$ 0.775
        & 0.09644 $\pm$ 0.01436 \\

        Base RAG
        & 1
        & --
        & --
        & 1.837 $\pm$ 1.378
        & 12.935 $\pm$ 5.168
        & 86.060 $\pm$ 0.862
        & 0.05099 $\pm$ 0.01097 \\

        \midrule

        TRACE (reference)
        & 2
        & 3
        & 0.02905 $\pm$ 0.00680
        & 2.710 $\pm$ 3.297
        & 13.414 $\pm$ 4.892
        & 86.601 $\pm$ 0.920
        & \textbf{0.04349 $\pm$ 0.01051} \\

        TRACE (more revisions)
        & 3
        & 3
        & 0.02905 $\pm$ 0.00680
        & \textbf{2.921 $\pm$ 3.297}
        & \textbf{14.332 $\pm$ 6.009}
        & 86.650 $\pm$ 0.863
        & 0.04415 $\pm$ 0.01023 \\

        TRACE (short window)
        & 2
        & 2
        & 0.02916 $\pm$ 0.00681
        & 2.828 $\pm$ 3.354
        & 14.256 $\pm$ 6.132
        & \textbf{86.826 $\pm$ 0.936}
        & 0.04571 $\pm$ 0.01193 \\

        TRACE (long window)
        & 2
        & 5
        & \textbf{0.02857 $\pm$ 0.00654}
        & 2.778 $\pm$ 3.335
        & 13.753 $\pm$ 6.625
        & 86.764 $\pm$ 0.767
        & 0.04561 $\pm$ 0.01262 \\

        \bottomrule
    \end{tabular}

    \begin{tablenotes}[flushleft]
        \footnotesize

        \item
        Bold denotes the best mean in each metric column,
        not statistical significance.
        $K$ is the maximum generation budget, including
        the initial draft; KEEP consumes a decision slot
        without generating another reply.
        $W$ denotes the trajectory window.
        Retrieval uses up to three examples.

        \item
        Both MSE metrics use 20 fixed normalized style features.
        State Prediction MSE compares the pre-generation forecast
        with the observed reference state, whereas Reply Style MSE
        compares the generated reply with the reference reply.
        Baselines have no state predictor (--).
        BLEU is computed at the corpus level within each user;
        ROUGE-L and BERTScore are F1 scores multiplied by 100.

    \end{tablenotes}

    \end{threeparttable}
\end{table*}

RAG improves all four reply-level metric means over Base LLM.
The reference TRACE configuration achieves further improvements,
increasing BLEU from 1.837 to 2.710 and BERTScore from 86.060
to 86.601 relative to RAG.
It also reduces Reply Style MSE from 0.05099 to 0.04349,
a relative reduction of 14.72\%.
These results indicate closer agreement with the reference
replies under the evaluated automatic metrics.
The comparison reflects the combined effects of explicit
profile information and the generation procedure;
individual component contributions are examined in
Section~\ref{sec:ablation-study}.

\subsubsection{State Prediction and Reply Quality}
\label{sec:state-prediction-analysis}

At a fixed generation budget of $K=2$, increasing $W$ from
2 to 3 and 5 reduces mean State Prediction MSE from
0.02916 to 0.02905 and 0.02857, respectively.
Among the three windows evaluated in the main experiment,
$W=5$ has the lowest forecast error, reducing it by 1.64\%
relative to the reference setting.
However, this improvement does not translate into lower
final reply-style error:
Reply Style MSE increases from 0.04349 at $W=3$ to
0.04561 at $W=5$, a relative increase of 4.88\%.
Meanwhile, $W=2$ achieves the highest BERTScore in the main
comparison (86.826), despite having the highest forecast
error among these three windows.
Forecast accuracy alone therefore does not identify the
configuration with the best downstream reply metrics.

Increasing $K$ from 2 to 3 at $W=3$ leaves State Prediction
MSE unchanged because both configurations share the same
pre-generation forecast.
The larger budget yields the highest BLEU and ROUGE-L
in the main comparison, but increases Reply Style MSE
from 0.04349 to 0.04415.
Thus, the reference configuration provides the closest
style-feature match among the four TRACE settings,
while other configurations favor lexical overlap or
BERTScore.
These mean comparisons are descriptive; the reported
between-user standard deviations do not by themselves
establish pairwise statistical significance.

\subsection{Ablation Study}
\label{sec:ablation-study}
Table~\ref{tab:trace-ablation} reports the interventions defined
in Section~\ref{sec:ablation-setup}.
All comparisons use the Full TRACE reference in this table.

\begin{table*}[t]
\centering
\caption{
Ablation results on 216 test emails from eight users.
Values are equally weighted user means $\pm$ between-user
sample standard deviations, with one online trajectory per user
and variant. Exact matching requests may share saved responses.
Bold indicates the best mean, not statistical significance.
}
\label{tab:trace-ablation}
\footnotesize
\setlength{\tabcolsep}{3pt}
\renewcommand{\arraystretch}{1.1}

\resizebox{\textwidth}{!}{%
\begin{tabular}{lccccc}
\toprule
Method
& \shortstack{State Prediction\\MSE $\downarrow$}
& BLEU $\uparrow$
& ROUGE-L $\uparrow$
& BERTScore $\uparrow$
& \shortstack{Reply Style\\MSE $\downarrow$} \\
\midrule

Full TRACE
& 0.02905 $\pm$ 0.00680
& \textbf{2.999 $\pm$ 3.274}
& 14.532 $\pm$ 5.673
& 86.804 $\pm$ 0.871
& \textbf{0.04195 $\pm$ 0.01008} \\

\midrule

\shortstack[l]{w/o Global Lexical\\Summary}
& 0.02905 $\pm$ 0.00680
& 2.468 $\pm$ 3.097
& 13.460 $\pm$ 5.592
& 86.635 $\pm$ 0.771
& 0.04475 $\pm$ 0.01006 \\

\shortstack[l]{Shuffled Contact\\TF-IDF}
& 0.02905 $\pm$ 0.00680
& 2.699 $\pm$ 2.915
& 14.391 $\pm$ 4.645
& 86.837 $\pm$ 0.833
& 0.04234 $\pm$ 0.00920 \\

w/o Temporal Modeling
& \textbf{0.02787 $\pm$ 0.00662}
& 2.903 $\pm$ 3.317
& \textbf{14.964 $\pm$ 5.481}
& \textbf{86.899 $\pm$ 0.766}
& 0.04211 $\pm$ 0.01137 \\

w/o Q-learning
& 0.02905 $\pm$ 0.00680
& 2.996 $\pm$ 3.238
& 14.200 $\pm$ 5.959
& 86.707 $\pm$ 0.967
& 0.04293 $\pm$ 0.01038 \\

w/o Revision
& 0.02905 $\pm$ 0.00680
& 2.973 $\pm$ 3.300
& 14.563 $\pm$ 5.630
& 86.766 $\pm$ 0.955
& 0.04269 $\pm$ 0.01070 \\

\bottomrule
\end{tabular}%
}
\end{table*}

Full TRACE achieves the highest mean BLEU (2.999) and the lowest
mean Reply Style MSE (0.04195).
Removing the explicit global lexical summary lowers BLEU,
ROUGE-L, and BERTScore and increases Reply Style MSE by $6.66\%$.
This comparison concerns the explicit summary, since the
relationship contrast still uses cross-contact statistics.
Shuffling contact TF-IDF, removing Q-learning, and removing
revision increase mean Reply Style MSE by $0.91\%$, $2.33\%$,
and $1.76\%$, respectively.
These four interventions leave the forecaster and its historical
inputs unchanged, explaining their identical State Prediction MSE.

Removing temporal modeling reduces State Prediction MSE from
0.02905 to 0.02787 ($4.06\%$) and yields the highest mean ROUGE-L
and BERTScore.
However, BLEU decreases from 2.999 to 2.903, and Reply Style MSE
increases slightly from 0.04195 to 0.04211.
With retrieval matched throughout training and testing, the
lower forecast error therefore does not translate into lower
final reply-style error in this comparison.
The temporal component has mixed observed effects across metrics.

For the primary endpoint, Reply Style MSE, none of the five
two-sided exact paired-user sign-flip tests is significant after
Holm correction across the five comparisons
\cite{holm1979simple} (minimum adjusted $p=0.15625$).
The observed differences therefore do not establish reliable
component-level improvements in Reply Style MSE.
Given the eight-user sample, one online trajectory per variant,
shared responses where requests match exactly, and the previously
observed test set, we treat these findings as exploratory.

\raggedbottom
\subsection{Sensitivity to $K$ and $W$}
\label{sec:kw-sensitivity}

We evaluate all 30 combinations of
$K\in\{1,2,3,4,5\}$ and $W\in\{2,3,4,5,8,12\}$.
Figure~\ref{fig:trace-kw-sensitivity} in
Appendix~\ref{app:kw-sensitivity} reports equally
weighted user means on 65 development emails from
the same eight users. For each user, we select the
latest $\max(1,\lfloor0.2N_u\rfloor)$ eligible training
cases, ordered by incoming-message time and case
identifier; $N_u$ counts cases whose replies arrive
before that user's first test query. Replay warms the
policy chronologically, and no original test case
enters this analysis. Development scores are therefore
not directly comparable with Table~\ref{tab:trace-six-methods}.

State Prediction MSE is lowest at $W=12$ (0.02476),
but only 39.24\% of development cases use temporal
fitting at that window; the rest use historical means.
The best observed BLEU, ROUGE-L, BERTScore, and
Reply Style MSE occur at $(K,W)=(5,2)$, $(5,5)$,
$(1,3)$, and $(3,8)$, respectively. Thus, better
forecasts or larger revision budgets do not
consistently improve reply quality. These optima
are descriptive: each user--configuration pair
has one online trajectory, and identical requests
may reuse cached responses.
\section{Conclusion}
\label{sec:conclusion}

We introduced relationship-specific continual reply prediction
for one-to-many private communication and presented TRACE,
which combines shared lexical habits, contact-specific
preferences, temporal state estimates, and a feedback-driven
revision policy around a frozen language model.

On 216 Enron test emails from eight users, TRACE achieves better
mean reply-level metrics than Base LLM and Base RAG.
The ablations do not establish statistically reliable
component-level improvements in Reply Style MSE after
multiple-comparison correction. More accurate forecasts and
larger revision budgets do not consistently improve reply-level
metrics. These exploratory findings motivate evaluation
with broader correspondence data and independent human judgments.
\section{Limitations}
\label{sec:limitations}

Our evaluation uses a curated Enron subset containing eight
users and 35 recurring relationships, together with a single
language model. This setting provides limited evidence about
generalization to other communication domains, languages,
or relationships with little historical data. Each
user--configuration pair contributes only one online
trajectory, and some generation responses are shared through
exact-request cache reuse. The small user sample, stochastic
generation, and prior inspection of the test set limit the
strength of statistical conclusions. Our findings should
therefore be interpreted as exploratory.

The evaluation relies on automatic comparisons with a single
observed reply. BLEU, ROUGE-L, and BERTScore measure reference
similarity, while the two MSE metrics capture only 20 predefined
style features. These measures cannot fully represent
communicative appropriateness or an author's preferences,
and alternative replies may be equally suitable. The revision
policy also receives rewards based on Reply Style MSE,
so improvements on this objective require independent
validation through human judgments. We do not evaluate
author satisfaction, factual reliability, or reductions
in editing effort through a user study.

Temporal estimates are inferred from limited observations
and may conflate changes in expression with changes in topic
or task. Insufficient history triggers a fallback to
historical means, which complicates the interpretation of
window-size comparisons. Moreover, the historical-mean
ablation continues to incorporate newly observed replies;
it therefore tests temporal forecasting rather than the
benefit of continual updating itself. The ablations do not
establish statistically reliable component-level improvements
in Reply Style MSE after multiple-comparison correction.

Finally, our protocol replays recorded correspondence and
assumes that actual replies become available as feedback.
For interactions with tied timestamps, prediction precedes
feedback by convention in our replay protocol; the recorded
timestamps alone do not establish the finer-grained
real-world ordering of these events.
The protocol does not capture how authors would respond to
generated suggestions or how assistant use might change
subsequent communication. The setting concerns private
correspondence, but the current API-based implementation
does not provide a formal privacy guarantee.

\bibliography{references}

\appendix
\appendix

\section*{Appendix}

\section{Communication Network}
\label{app:communication_network}

Figure~\ref{fig:enron-dataset} shows the retained
user--contact relationships within the broader
Enron communication network. Additional neighboring
accounts provide context and are not part of
the study subset.

\section{Sensitivity Curves for $K$ and $W$}
\label{app:kw-sensitivity}

\begin{figure*}[p]
\centering

\includegraphics[
    width=0.82\textwidth,
    height=0.34\textheight,
    keepaspectratio
]{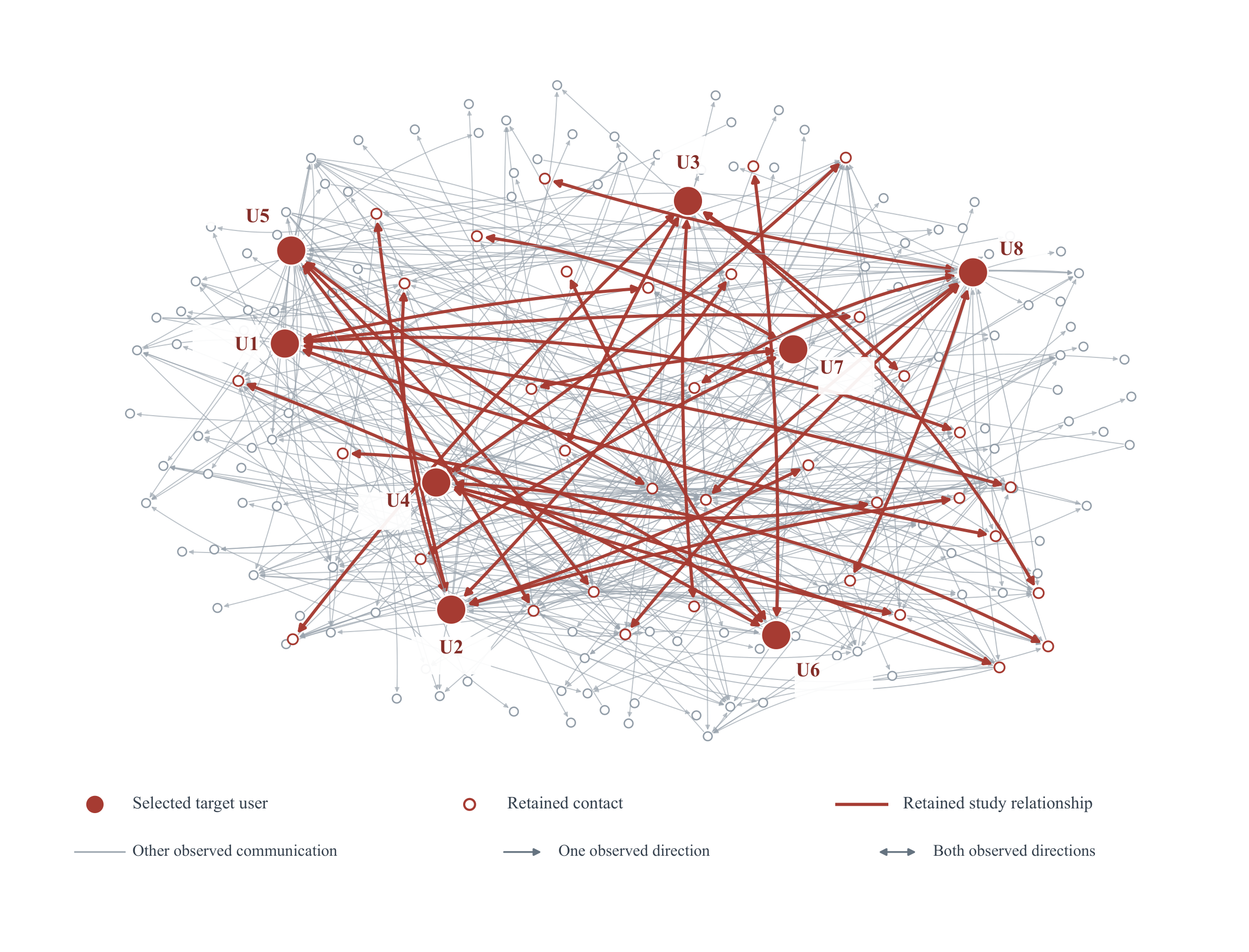}

\caption{
    Communication context of the curated Enron subset:
    43 study accounts and 120 neighboring accounts.
    Filled red nodes denote the eight target users,
    open red nodes denote retained contacts,
    and thick red links denote the 35 retained
    user--contact relationships.
    Gray links show other observed communication.
    Arrowheads indicate observed directionality;
    node positions are schematic.
}
\label{fig:enron-dataset}

\medskip

\includegraphics[
    width=\textwidth,
    height=0.38\textheight,
    keepaspectratio
]{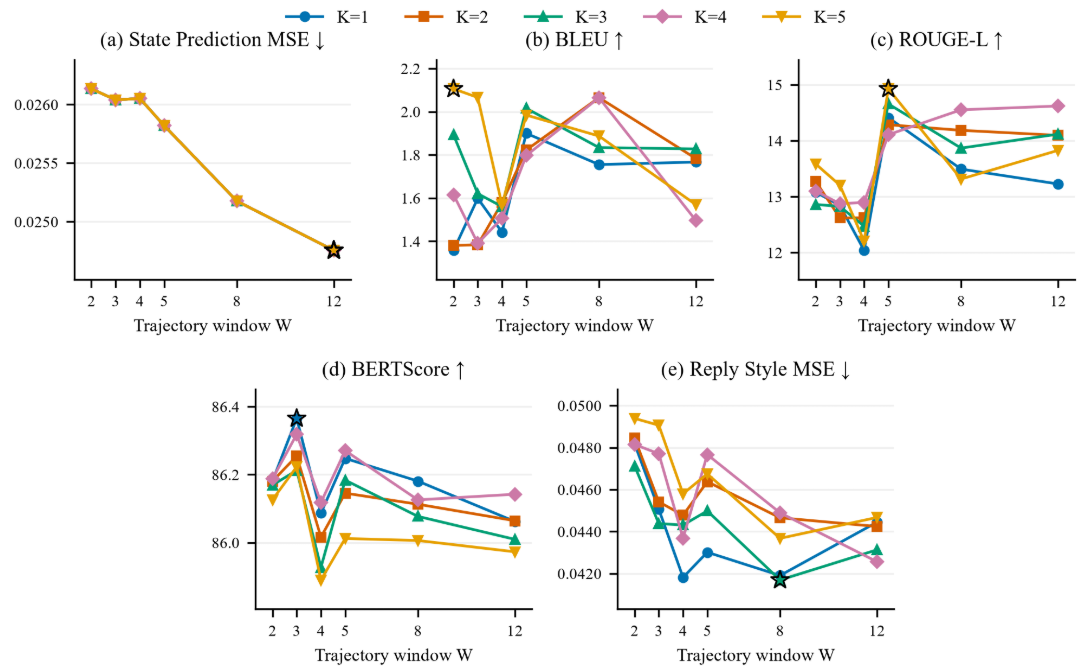}

\caption{
    Sensitivity to $W$ at five generation budgets $K$
    on 65 development emails from eight users.
    Points are equally weighted user means;
    lines connect observations without smoothing.
    $K$ includes the initial draft; $W$ is distinct
    from the retrieval count.
    Stars mark the best observed means, including ties,
    rather than statistical significance.
    State Prediction MSE curves overlap because
    forecasting is independent of $K$.
    Lower is better for MSE; higher is better
    for BLEU, ROUGE-L, and BERTScore.
}
\label{fig:trace-kw-sensitivity}

\end{figure*}

Figure~\ref{fig:trace-kw-sensitivity} presents
the sensitivity curves for the maximum generation
budget $K$ and trajectory window $W$ on the
development subset.

\section{Data Provenance and Preprocessing}
\label{app:source-provenance}

We obtained the Enron Email-Reply Dataset by
\citet{ilie2025enronreply} through its Kaggle distribution%
\footnote{\url{https://www.kaggle.com/datasets/oanannv/enron-email-reply-dataset}}.
The corresponding Zenodo record was released on August 13, 2025.
The public release contains 15,377 message--reply pairs.

Our downstream processing normalizes the case and surrounding
whitespace of participant addresses and parses timestamps for
ordering and grouping. We purposively selected eight users with
multiple contacts and frequent correspondence, retaining 35
user--contact relationships and 1,020 pairs. The selected incoming
and reference texts match their source CSV fields exactly.
We retain the supplied message--reply associations rather than
reconstructing pairs from raw emails. Timestamp handling and
the train/test split follow Section~5.2.

According to the release documentation, the publisher matched
messages using subjects, participants, and chronology, with a
maximum three-day reply gap. The publisher also reports removing
quoted content, HTML, and attachment markers, expanding
contractions, and normalizing whitespace. These are upstream
procedures reported by the publisher; our experiments use the
resulting preprocessed texts.

Both downloaded release archives include a \texttt{LICENSE}
file specifying CC BY 4.0, consistent with the Zenodo metadata.
The Kaggle page's top-level license metadata instead lists MIT;
we report the archive license explicitly and retain attribution
to the dataset creator.

\section{Style Feature Specification}
\label{app:trace-style-features}

The 20 style features are token count, sentence count,
mean sentence length, mean word length, paragraph count,
moving-average type--token ratio, function-word ratio,
long-word ratio, contraction rate, first-person singular
rate, first-person plural rate, second-person rate,
greeting, direct address, closing, signature,
modal-verb rate, intensifier rate, question-mark rate,
and exclamation-mark rate. Rates count occurrences
per 100 lexical tokens; greeting, direct address,
closing, and signature are binary indicators.

Token count, sentence count, mean sentence length,
and paragraph count use $\log(1+x)/\log(1+s)$,
with $s=500,20,80,10$, respectively.
Other coordinates use $x/s$: $s=12$ for mean word
length, $30$ for the three pronoun rates,
$20$ for the other rates, and $1$ for ratios and
binary indicators. All coordinates are clipped to
$[0,1]$ and receive equal weight in MSE.
The scales are fixed rather than fitted to the data.

The controller uses 11 of these coordinates:
token count, sentence count, mean sentence length,
the three pronoun rates, contraction rate,
modal-verb rate, question-mark rate, greeting,
and direct address. Rewards and final style
evaluation retain all 20 coordinates.

\end{document}